\documentclass[]{rsos}

\usepackage{float}
\usepackage{nameref}

\begin{document}

\title{The Minor Fall, the Major Lift: Inferring Emotional Valence of Musical Chords through Lyrics}

\title{The Minor Fall, the Major Lift: Inferring Emotional Valence of Musical Chords through Lyrics}

\author{
Artemy Kolchinsky$^{1,2,+}$,
Nakul Dhande$^{1,3+}$,
Kengjeun Park$^{1,4}$,
Yong-Yeol Ahn$^{1}$}

\address{$^{1}$ Department of Informatics, Indiana University, Bloomington, IN 47408, United States\\
$^{2}$ Santa Fe Institute, Santa Fe, NM 87501, United States (current affiliation)\\
$^{3}$ Amazon.com \\
$^{4}$ Epic.com \\
$^{+}$ these authors contributed equally to this work
}


\corres{Yong-Yeol Ahn\\
\email{yyahn@iu.edu}}

\begin{abstract}
We investigate the association between musical chords and lyrics by analyzing a large dataset of user-contributed guitar tablatures. Motivated by the idea that the emotional content of chords is reflected in the words used in corresponding lyrics, we analyze associations between lyrics and chord categories. We also examine the usage patterns of chords and lyrics in different musical genres, historical eras, and geographical regions. Our overall results confirms a previously known association between Major chords and positive valence. We also report a wide variation in this association across regions, genres, and eras. Our results suggest possible existence of different emotional associations for other types of chords.  
\end{abstract}

\begin{fmtext}

\section{Introduction}

The power of music to evoke strong feelings has long been admired and
explored~\cite{meyer-emotion-1961, dainow1977physical, garrido2011individual,
cross-music-2006, zentner2008emotions}.  Although music has accompanied humanity
since the dawn of culture~\cite{wallin2001origins} and its underlying
mathematical structure has been studied for many years~\cite{fauvel2003music,
tymoczko2006geometry, callender2008voice, helmholtz2009sensations},
understanding the link between music and emotion remains a
challenge~\cite{meyer-emotion-1961, hunter-emotion-2010, juslin2008emotional,
eerola-review-2013}.  

\end{fmtext}
\maketitle

The study of music perception has been dominated by
methods that \emph{directly} measure emotional responses, such as self-reports,
physiological and cognitive measurements, and developmental
observations~\cite{hunter-emotion-2010, eerola-review-2013}. Such
methods may produce high-quality data, but the data collection process involved is
both labor- and resource-intensive. As a result, creating large datasets and
discovering statistical regularities has been a challenge.

Meanwhile, the growth of music databases~\cite{IMSLP, lastfm, bertin2011msd,
huron2002humdrum, ultimateguitar, gracenote} as well as the advancement of the
field of Music Information Retrieval (MIR)~\cite{downie2008mirex,
kim2010review, yang-machine-2012} opened new avenues for data-driven studies
of music. For instance, sentiment analysis~\cite{pang-sentiment-2008,
dodds2011temporal, liu-sentiment-2012, gonccalves2013comparing} has been
applied to uncover a long-term trend of declining valence in popular song
lyrics~\cite{dodds2010measuring, dewall2011tuning}. It has been shown that the
lexical features from lyrics~\cite{yang2004disambiguating, yang2008toward,
laurier2008multimodal,hu2009lyric, van2010automatic, mihalcea-lyrics-2012},
metadata~\cite{schuller2010determination}, social
tags~\cite{turnbull2009combining,bischoff2009music}, and audio-based features
can be used to predict the mood of a song. There has been also an attempt to
examine the associations between lyrics and individual chords using a machine
translation approach, which confirmed the notion that Major and Minor chords
are associated with happy and sad words
respectively~\cite{ohara-inferring-2012}. 

Here, we propose a novel method to study the associations between
chord types and emotional valence. In particular, we use sentiment analysis to analyze \emph{chord categories} (e.g. Major and Minor) and their associations with sentiment and
words across genres, regions, and eras. 

\begin{figure}[th]
\centering
\includegraphics[width=1\textwidth]{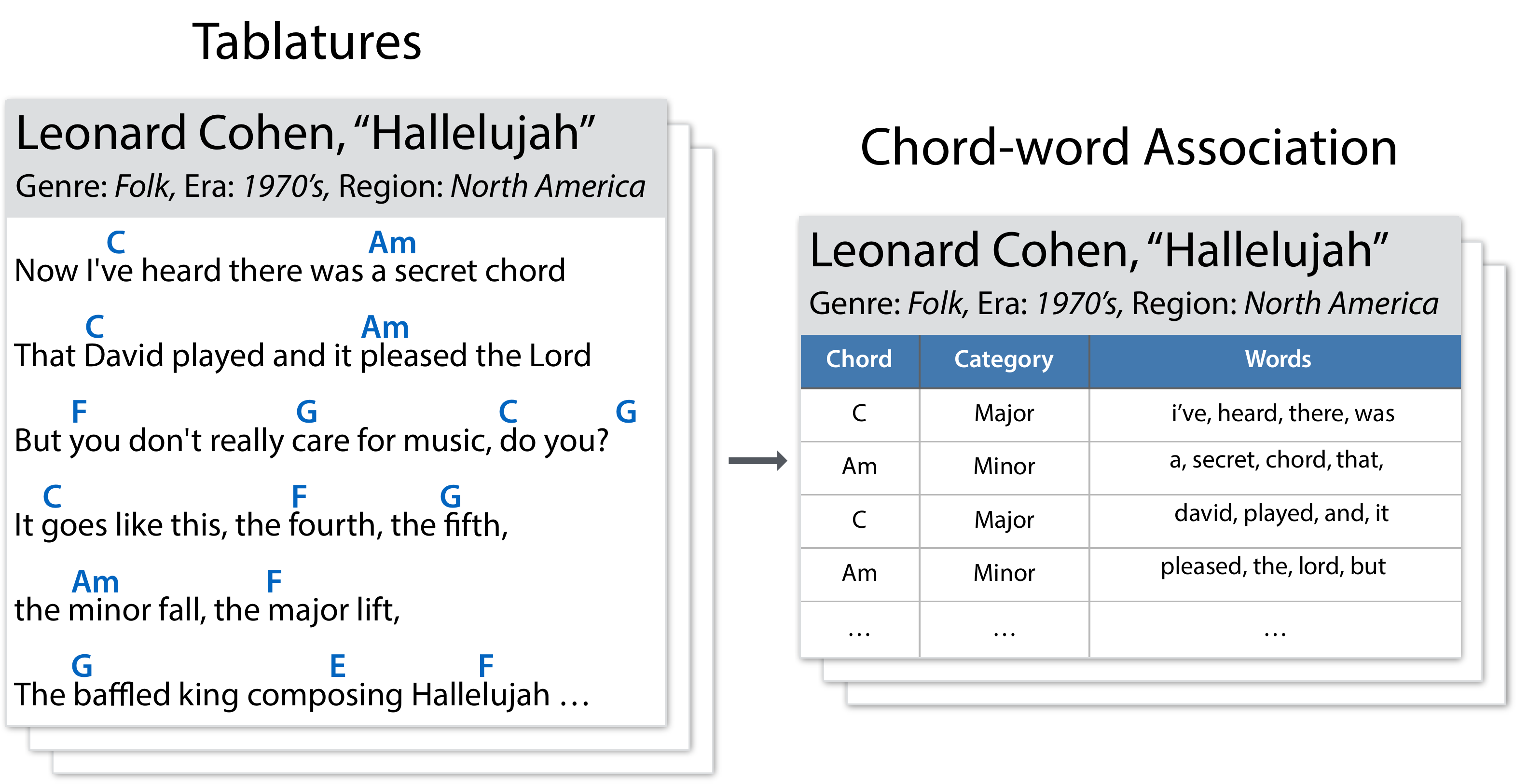}
\caption{A schematic diagram of our process of collecting guitar tablatures and
metadata, parsing chord-word associations, and analyzing the results using data
mining and sentiment analysis. Note that `Genre' is an album-specific label, while `Era', and `Region' are artist-specific labels (rather than song-specific).}
\label{fig:process}
\end{figure}

To do so, we adopt a sentiment analysis method that uses the crowd-sourced \emph{LabMT 1.0} valence lexicon~\cite{dodds2011temporal, bradley1999affective}. Valence is one of the two basic emotional axes~\cite{eerola-review-2013}, with higher valence corresponding to more attractive/positive emotion.
The lexicon contains valence scores ranging from $0.0$ (saddest) to $9.0$ (happiest) for $10,222$ common English words obtained by surveying Amazon's Mechanical Turk workers.
The overall valence of a piece of text, such as a sentence or document, is measured by averaging the valence score of individual words within the text. This method has been successfully used to obtain insight into a wide variety of corpora~\cite{bliss2012twitter,
mitchell2013geography, kloumann2012positivity, frank2013happiness}.

Here, we apply this sentiment analysis method to a dataset of guitar tablatures --- which contain both lyrics and chords --- extracted from \url{ultimate-guitar.com}. We collect all words that appear with a specific chord and create a large ``bag of words'' --- a frequency list of words --- for each chord (see
Fig.~\ref{fig:process}).  We perform our analysis by aggregating chords based on their `chord category' (e.g., Major chords, Minor chords, Dominant 7th chords, etc.). In addition, we also acquire metadata from the Gracenote API regarding the genre of albums, as well as era and geographical region of musical artists. We then perform our analysis of associations between lyrics sentiment and chords within the categories of genre, era, and region. Details of our methodology are described in the next section.

\section{Materials and Methods}\label{sec:methods}

Guitar tabs were obtained from
{\url{ultimate-guitar.com}}~\cite{ultimateguitar}, a large online
user-generated database of tabs, while information about album genre, artist era, and artist region was obtained from Gracenote API~\cite{gracenote}, an online musical metadata service.

\subsection{Chords-lyrics association}

\texttt{ultimate-guitar.com} is one of the largest user-contributed tab archives,
hosting more than 800,000 songs. We examined 123,837 songs that passed the following criteria: (1) we only kept guitar tabs and ignored those for other
instruments such as the ukulele; (2) we ignored non-English songs (those having less than half of their words in an English word list~\cite{SIL-International:Online} or identified as non-English by the \texttt{langdetect} library~\cite{langdetect}); (3) when multiple tabs were available for a song, we kept only the one with the highest user-assigned rating.
We then cleaned the raw HTML sources and extracted chords and lyrics transcriptions.
As an example, Fig.~\ref{fig:process} shows how the tablature of Leonard Cohen's ``Hallelujah''~\cite{HallelujahTab} is processed to produce a chord-lyrics table.

Sometimes, chord symbols appeared in the middle of words; in such cases, we associated the entire words with the chord that appears in its middle, rather than the previous chord.  
In addition, chords that could not be successfully parsed or that had no associated lyrics were dropped.

\subsection{Metadata collection using Gracenote API}
\label{methods-metadata}

We used the Gracenote API (\url{http://gracenote.com}) to obtain metadata about artists and song albums. We queried the title and the artist name of the 124,101 songs that were initially obtained from \texttt{ultimate-guitar.com}, successfully retrieving Gracenote records for 89,821 songs. Songs that did not match a Gracenote record were dropped.  For each song, we extracted the following metadata fields:
\begin{itemize}

\item The geographic \textbf{region} from which the artist originated (e.g. \emph{North America}).  This was extracted from the highest-level geographic labels provided by Gracenote.

\item The musical \textbf{genre} (e.g. \emph{50's Rock}). This was extracted from the second-level genre labels assigned to each album by Gracenote.

\item The historical \textbf{era}  at the level of decades (e.g. \emph{1970's}).  This was extracted from the first-level era labels assigned to each artist by Gracenote. Approximately 6,000 songs were not assigned to an era, in which case they were assigned to the decade of the album release year as specified by Gracenote.

\end{itemize}

In our analysis, we reported individual statistics only for the most popular regions (\emph{Asia}, \emph{Australia/Oceania}, \emph{North America}, \emph{Scandinavia}, \emph{Western Europe}), genres (top 20 most popular genres), and eras (1950's through 2010's).

\subsection{Determining chord categories}
\label{methods-chord-cats}

We normalized chord names and classified them into chord categories according to chord notation rules from an online resource~\cite{hakwright:Online}.  All valid chord names begin with one or two characters indicating the root note (e.g. \texttt{G} or \texttt{Bb}) which are followed by characters which indicate the \emph{chord category}.  We considered the following chord categories~\cite{benward2007music}:

\begin{itemize}

\item \textbf{Major}: A Major chord is a triad with a root, a major third and a perfect fifth. Major chords are indicated using either only the root note, or the root note followed by \texttt{M} or \texttt{maj}. For instance,  \texttt{F},
\texttt{FM}, \texttt{G}, \texttt{Gmaj} were considered Major chords.

\item \textbf{Minor}: A Minor chord is also a triad, containing a root, minor third and a perfect fifth. The notation for Minor chords is to have the root note followed by \texttt{m} or \texttt{min}.
For example, \texttt{Emin}, \texttt{F\#m} and \texttt{Bbm} were considered Minor chords. 

\item \textbf{7th}: A seventh chord has seventh interval in addition to a Major or
Minor triad. A \textbf{Major 7th} consists
of a Major triad and an additional Major seventh, and is 
indicated by the root note followed by \texttt{M7} or \texttt{maj7} (e.g. 
\texttt{GM7}). A \textbf{Minor 7th} consists of a Minor triad with an additional Minor seventh 
makes, and is indicated by the root note followed by 
\texttt{m7} or \texttt{min7} (e.g. \texttt{Fm7}). A \textbf{Dominant 7th} is a 
diatonic seventh chord that consists of a Major triad with 
additional Minor seventh, and is indicated by the root note followed by the 
numeral \texttt{7} or \texttt{dom7} (e.g. \texttt{D7}, \texttt{Gdom7}).

\item \textbf{Special chords with `*'}: In tab notation, the asterisk `*' is used to indicate special instructions and can have many different meanings. For instance, \texttt{G*} may indicate that the \texttt{G} should be played with a palm mute, with a single strum, or some other special instruction usually indicated in free text in the tablature. Because in most cases the underlying chord is still played, in this study we map chords with asterisks to their respective non-asterisk versions. For instance, we consider \texttt{G*} to be the same as \texttt{G} and \texttt{C7*} to be the same as \texttt{C7}. 

\item \textbf{Other chords}: 
There were several other categories of chords that we do not analyze individually in this study.  One of these is \textbf{Power} chords, which are dyads consisting of a root and a perfect fifth.  Because Power chords are highly genre specific, and because they sometimes function musically as Minor or Major chords, we eliminated them from our study.  For reasons of simplicity and statistical significance, we also eliminated several other categories of chords, such as \textbf{Augmented} and \textbf{Diminished} chords, which appeared infrequently in our dataset.
\end{itemize}

\begin{figure}[th]
	\centering
	\includegraphics[width=0.35\textwidth]{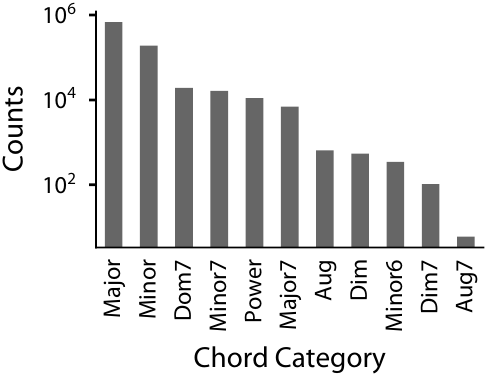}
\caption{
Prevalence of different chord categories within the dataset (note logarithmic scale).
}
\label{fig:category_prevalence}
\end{figure}

In total, we analyzed 924,418 chords (see the next subsection). Figure~\ref{fig:category_prevalence} shows the prevalence of different chord categories among these chords.

\subsection{Sentiment analysis}

Sentiment analysis was used to measure the valence (happiness vs. unhappiness) of chord lyrics.   We employed a simple methodology based on a crowd-sourced valence lexicon (\texttt{LabMT 1.0})~\cite{dodds2011temporal, bradley1999affective}. This method was chosen because (1) it is simple and scalable (2) it is \emph{transparent}, allowing us to calculate the contribution from each word to the final valence score, and (3) it has been shown to be useful in many
studies~\cite{bliss2012twitter, mitchell2013geography, kloumann2012positivity,
frank2013happiness}.  The \texttt{LabMT 1.0} lexicon contains valence scores ranging from 0.0 (saddest) to 9.0 (happiest) for 10,222 common English words as obtained by surveying Amazon's Mechanical Turk workers.  The valence assigned to some sequence of words (e.g. words in the lyrics corresponding to Major chords) was computed by mapping each word to its corresponding valence score and then computing the mean.  Words not found in the LabMT lexicon were ignored; in addition, following recommended practices for increasing sentiment signal~\cite{dodds2011temporal}, we ignored emotionally-neutral words having a valence strictly between 3.0 and 7.0. Chords that were not associated with any sentiment-mapped words were ignored.  The final dataset contained 924,418 chords from 86,627 songs.

\subsection{Word shift graphs}
\label{sec:methods-wordshift}

In order to show \emph{how} a set of lyrics (e.g. lyrics corresponding to songs in the \emph{Punk} genre) differs from the overall lyrics dataset, we use the \emph{word shift graphs}~\cite{dodds2010measuring,mitchell2013geography}. We designate the whole dataset as the \emph{reference} (baseline) corpus and call the set of lyrics that we want to compare \emph{comparison corpus}. The difference in their overall valence can now be broken down into the contribution from each individual word. Increased valence can result from either having a 
higher \emph{prevalence} (frequency of occurrence) of high-valence words
or a lower prevalence of low-valence words. Conversely, 
lower valence can result from having a higher prevalence of low-valence
words or a lower prevalence of high-valence words.  The percentage contribution of 
an individual word $i$ to the valence difference between a comparison and reference corpus can be expressed as: 
\begin{align*}
100\cdot\frac{\overbrace{\left(h_i-h^{\textrm{(ref)}}\right)}^{+/-}\overbrace{\left(p_{i}-p_{i}^{\textrm{(ref)}}\right)}^{\uparrow/\downarrow}}{\left|h^{\textrm{(comp)}}-h^{\textrm{(ref)}}\right|}
\end{align*} 
where $h_i$ is the valence score of word $i$ in the 
lexicon, $h^{\textrm{(ref)}}$ and $h^{\textrm{(comp)}}$ are 
the mean valences of the words in the reference corpus and 
comparison corpus respectively, $p_{i}$ is the normalized 
frequency of word $i$ in the comparison corpus, and
$p_{i}^{\textrm{(ref)}}$ is the normalized frequency of 
word $i$ in the reference corpus (normalized frequencies 
are computed as $p_i = \frac {n_{i}}{\sum_{i'}{n_{i'}}}$, 
where $ n_i $ is the number of occurrences of a word $i$).
The first term (indicated by `+/-') measures the difference 
in word valence between word $i$ and the mean valence of the reference corpus,
while the second term (indicated by $\uparrow/\downarrow$)
looks at the difference in word prevalence between the 
comparison and reference corpus.  In plotting the word shift
graphs, for each word we use +/- signs and blue/green bar colors to indicate the (positive or negative) sign of the valence term and 
$\uparrow/\downarrow$ arrows to indicate the sign of the prevalence term.

\subsection{Model comparison}

In the Results section, we evaluate what explanatory factors (chord category, genre, era, and region) best account for differences in valence scores. Using the \texttt{statsmodels} toolbox~\cite{statsmodels}, we estimated linear regression models where the mean valence of each chord served as the response variable and the most popular chord categories, genres, eras, and regions served as the categorical predictor variables.  The variance of the residuals was used to compute the proportion of variance explained when using each factor in turn.

We also compared models that used combinations of factors.  As before, we fit linear models that predicted valence.  Now, however, explanatory factors were added in a greedy fashion, with each additional factor to minimize the Aikake information criterion (AIC) of the overall model.

\section{Results}

\subsection{Valence of chord categories}

\begin{figure}[th]
	\centering
	\includegraphics[width=\textwidth]{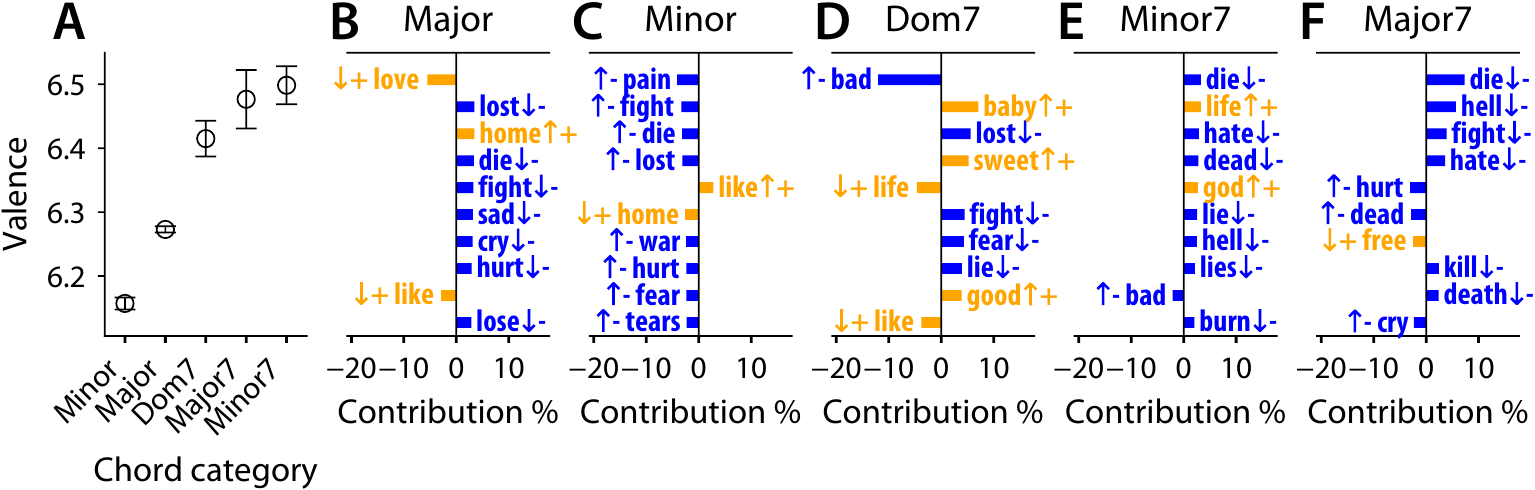}
\caption{(A): Mean valence for chord categories. Error bars indicate 95\% CI of
the mean (error bars for Minor and Major chords are smaller than the symbols).
(B-F): Word shift plots for different chord categories.	 High-valence words are
indicated using `+' symbol and orange color, while low-valence words are
indicated by `-' symbol and blue color.  Words that are overexpressed in the
lyrics corresponding to a given chord category are indicated by `$\uparrow$',
while underexpressed words are indicated by `$\downarrow$'.
}
\label{fig:category_valence}
\end{figure}

We measure the mean valence of lyrics associated with different chord
categories (Figure~\ref{fig:category_valence}A). We find that major chords have higher
valence than Minor chords, concurring  with numerous studies which argue that
human subjects perceive Major chords as more emotionally positive than Minor
chords~\cite{crowder1984perception,kastner1990perception, hunter2010feelings}.
However, our results suggest that Major chords are not the happiest: all three
categories of 7th chords considered here (Minor 7th, Major 7th, and Dominant
7th) exhibit higher valence than Major chords.  This effect holds with high
significance ($p \ll 10^{-3}$ for all, one-tailed Mann-Whitney tests). 

In Fig.~\ref{fig:category_valence}B-F, we use the word shift graphs~\cite{dodds2010measuring} to identify words that contribute most to
the difference between the valence of each chord category and baseline (mean
valence of all lyrics in the dataset). For instance, ``lost'' is a
lower-valence word (blue color and `-' sign) that is underexpressed in Major
chords (`$\downarrow$' sign), increasing the mean valence of Major chords above baseline. Many negative words, such as ``pain'', ``fight'', ``die'', and ``lost'' are
overexpressed in Minor chord lyrics and underrepresented in Major chord lyrics.

Although the three types of 7th chords have similar valence scores
(Fig.~\ref{fig:category_valence}A), word shift graphs reveals that they may
have different ``meanings'' in terms of associated words.  Overexpressed
high-valence words for Dominant 7th chords include terms of endearment or affection, such as ``baby'', ``sweet'', and
``good'' while
for Minor 7th chords they are ``life'' and ``god''. Lyrics associated with Major 7th chords, on the
other hand, have a stronger under-representation of negative words (e.g. ``die'' and ``hell'').

\subsection{Genres}

\begin{figure}[th]
	\begin{center}
		\includegraphics[width=1\textwidth]{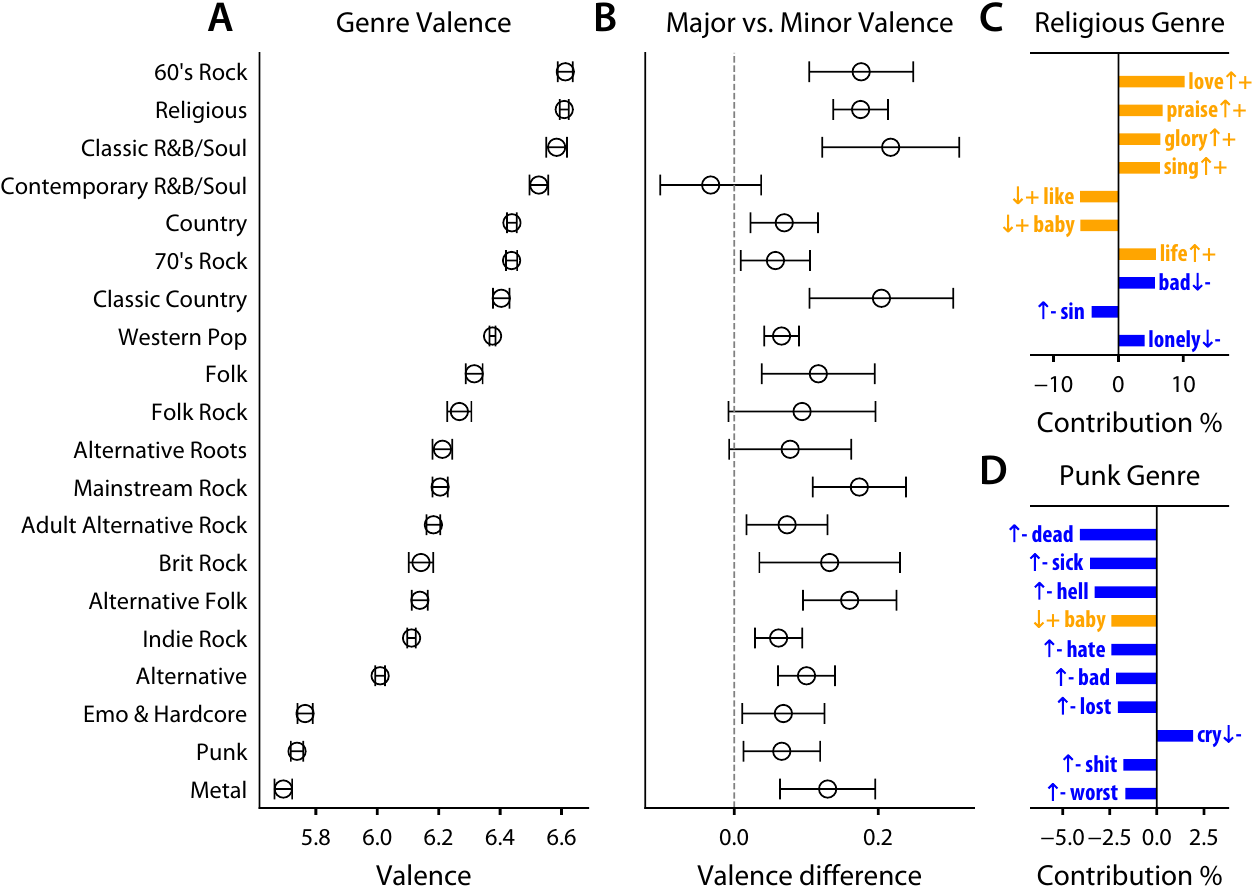}
	\end{center}
	\caption{(A) Mean valence of lyrics by album genre. (B) Major vs. Minor valence differences for lyrics by album genre. (C) Word shift plot for the \emph{Religious} genre. (D) Word shift plot for the \emph{Punk} genre.  See caption of Fig.~\ref{fig:category_valence} for explanation of word shift plot symbols.}
	\label{fig:genrefig}
\end{figure}

We analyze the emotional content and meaning of lyrics from albums in 
different musical genres.  Fig.~\ref{fig:genrefig}A shows the mean valence of
different genres, demonstrating that an intuitive ranking emerges when genres
are sorted by valence: \emph{Religious} and \emph{60's Rock} lyrics reside at
the positive end of the spectrum while \emph{Metal} and \emph{Punk} lyrics
appear at the most negative.

As mentioned in the previous section, Minor chords have a lower mean valence than
Major chords.  We computed the numerical \emph{differences} in valence between
Major and Minor chords for albums in different genres
(Fig.~\ref{fig:genrefig}B).  All considered genres, with the exception of
\emph{Contemporary R\&B/Soul}, had a mean valence of Major chords higher than
that of Minor chords.  Some of the genres with the largest Major vs. Minor
valence differences include \emph{Classic R\&B/Soul}, \emph{Classic Country},
\emph{Religious}, and \emph{60's Rock}.

We show word shift graphs for two musical genres: \emph{Religious} (Fig.~\ref{fig:genrefig}C) and \emph{Punk} (Fig.~\ref{fig:genrefig}D). The highest contributions to the \emph{Religious} genre come from the overexpression of high-valence words having to do with worship (``love'', ``praise'', ``glory'', ``sing''). Conversely, the highest contributions to the \emph{Punk} genre come from the overexpression of low-valence words (e.g. ``dead'', ``sick'', ``hell'').  Some exceptions exist: for example, \emph{Religious} lyrics underexpress high-valence words such as ``baby'' and ``like'', while \emph{Punk} lyrics underexpress the low-valence word ``cry''.

\subsection{Era}

\begin{figure}[th!]
	\begin{center}
		\includegraphics[width=\textwidth]{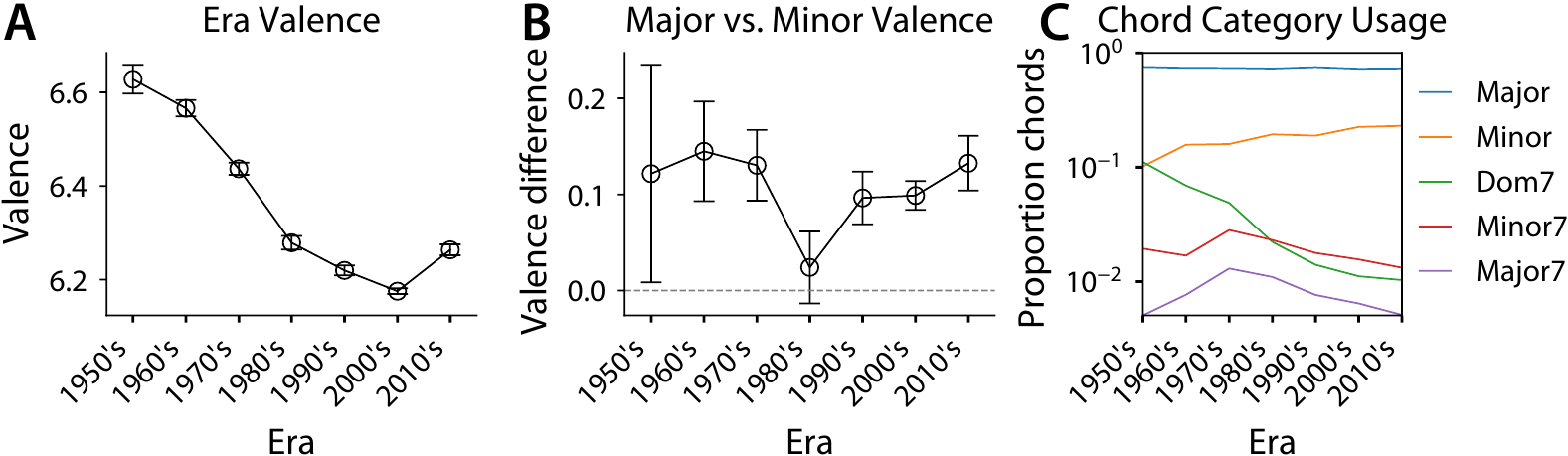}
	\end{center}
	\caption{(A) Mean valence of lyrics by artist era. (B) Major vs. Minor valence differences  by artist era. (C) Proportion of chords in each chord category in different eras (note logarithmic scale).
}
	\label{fig:erafig}
\end{figure}

In this section, we explore sentiment trends for artists across different historical eras.  Fig.~\ref{fig:erafig}A shows the mean valence of lyrics in different eras. We find that valence has steadily decreased since the 1950's, confirming results of previous sentiment analysis of lyrics~\cite{dodds2010measuring}, which attributed the decline to the recent emergence of `dark' genres such as metal and punk.  However, our results demonstrate that this trend has recently undergone a reversal: lyrics have higher valence in the 2010's era than in the 2000's era.

As in the last section, we analyze differences between Major and Minor chords for lyrics belonging to different eras (Fig.~\ref{fig:erafig}B).  Although Major chords have a  generally higher valence than Minor chords, surprisingly this distinction does not hold in the 1980's era, in which Minor and Major chord valences are similar.  The genres in the 1980's that had Minor chords with higher mean valence than Major chords --- in other words, which had an `inverted' Major/Minor valence pattern --- include \emph{Alternative Folk}, \emph{Indie Rock}, and \emph{Punk} (data not shown).

Finally, we report changes in chord usage patterns across time. Fig.~\ref{fig:erafig}C shows the proportion of chords belonging to each chord category in different eras (note the logarithmic scale). Since the 1950's, Major chord usage has been stable while Minor chords usage has been steadily growing. Dominant 7th chords have become less prevalent, while Major 7th 
and Minor 7th chords had an increase in usage during the 1970's.

The finding that Minor chords have become more prevalent while Dominant 7th
chords have become rarer agrees with a recent data-driven study
of the evolution of popular music genres~\cite{mauch2015evolution}.  The authors
attribute the latter effect to the decline in the popularity of blues and jazz,
which frequently use Dominant 7th chords.  However, we find that this effect
holds widely, with Dominant 7th chords diminishing in prevalence even when we
exclude genres associated with Blues and Jazz (data not shown).
More qualitatively, musicologists have argued that
 many popular music styles in the 1970's exhibited a decline in the use of
Dominant 7th chords and a growth in the use of Major 7th and Minor 7th
chords~\cite{stephenson2002listen} --- clearly seen in the corresponding increases in Fig.~\ref{fig:erafig}C.

\subsection{Region}

\begin{figure}[th]
	\begin{center}
		\includegraphics[width=0.75\textwidth]{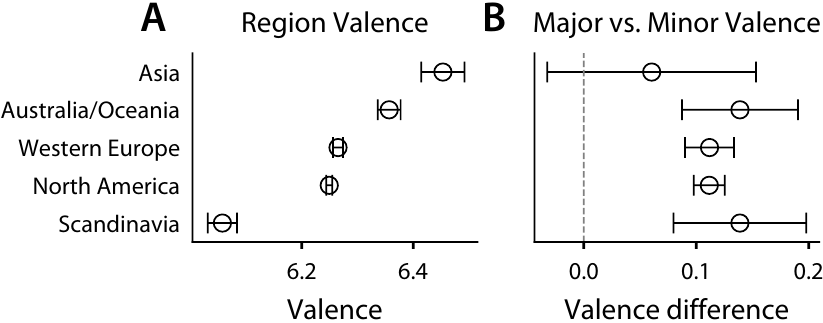}
	\end{center}
	\caption{(A) Mean valence of lyrics  by artist region. (B) Major vs. Minor valence differences by artist region.}
	\label{fig:regionfig}
\end{figure}

In this section, we evaluate the emotional content of lyrics from artists in different geographical regions. Fig.~\ref{fig:regionfig}A shows that artists from Asia have the highest-valence lyrics, followed by artists from \emph{Australia/Oceania}, \emph{Western Europe}, \emph{North America}, and finally \emph{Scandinavia}, the lowest valence geographical region.  The latter region's low valence is likely due to the over-representation of  `dark' genres (such as metal) among Scandinavian artists~\cite{scaruffi2009history}.

As in previous sections, we compare differences in valence of Major and Minor
chords for different regions (Fig.~\ref{fig:regionfig}B). All regions except
\emph{Asia} have a higher mean valence for Major chords than Minor chords, while
for the \emph{Asian} region there is no significant difference.

There are several important caveats to our geographical analysis.  In particular, our dataset consisted of only English-language songs, and is thus unlikely to be representative of overall musical trends in non-English speaking countries.  This bias, along with others, is discussed in more depth in the Discussion section.

\subsection{Model comparison}\label{sub:modelcomparison}

\begin{figure}[th]
	\begin{center}
		\includegraphics[width=0.65\textwidth]{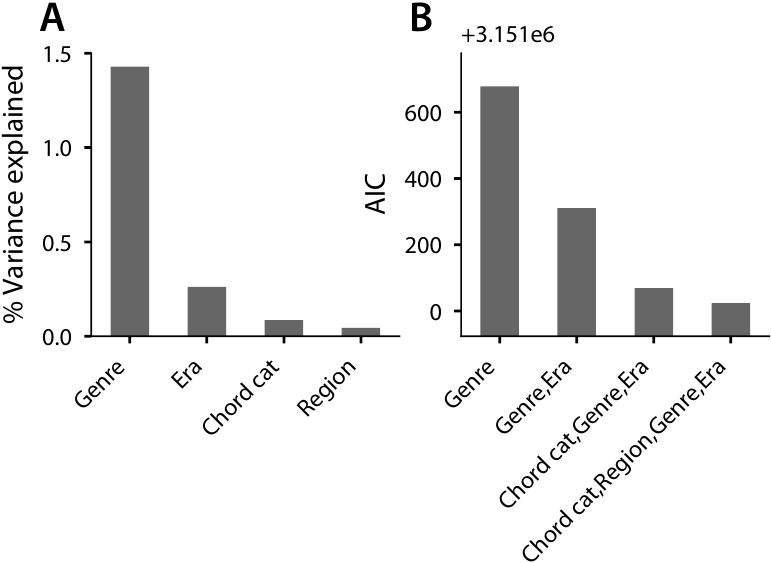}
	\end{center}
	\caption{(A) \% of explained valence variation when using chord category, genre, era, and region as categorical predictor variables. (B) AIC model selection scores for models that predict valence using different explanatory factors. The model that includes all explanatory factors is preferred.}
	\label{fig:modelcomparison}
\end{figure}

We have shown that mean valence varies as a function of chord category, genre, era, and region (which are here called \emph{explanatory factors}). We evaluate what explanatory factors best account for differences in valence scores. Fig.~\ref{fig:modelcomparison}A shows the proportion of variance explained when using each factor in turn as a predictor of valence.  This shows that genre explains most variation in valence, followed by era, chord category, and finally region.

It is possible that variation in valence associated with some explanatory factors is in fact `mediated' by other factors.  For example, we found that mean valence declined from the 1950's era through the 2000's, confirming previous work~\cite{dodds2010measuring} that explained this decline by the growing popularity of `dark' genres like \emph{Metal} and \emph{Punk} over time; this is an example in which valence variation over historical eras is argued to actually be attributable to variation in the popularities of different genres.  As another example, it is possible that Minor chords are lower valence than Major chords because they are overexpressed in dark genres, rather than due to their inherent emotional content.

We investigate this effect using statistical model selection. For instance, if the valence variation over chord categories can be entirely attributed to genre (i.e. darker genres have more Minor chords), then model selection should prefer a model that contains only the genre explanatory factor to the one that contains both the genre and chord category explanatory factors.

We fit increasingly large models while computing their Aikake information criterion (AIC) scores, a model selection score (lower is better).  As Fig.~\ref{fig:modelcomparison}B shows, the model that includes all four explanatory factors has the lowest AIC, suggesting that chord category, genre, era, and region are all important factors for explaining valence variation.  

\section{Discussion}

In this paper, we propose a novel data-driven method to uncover emotional valence associated with different chords as well as different geographical regions, historical eras, and musical genres.  We then apply it to a novel dataset of guitar tablatures extracted from \url{ultimate-guitar.com} along with musical metadata provided by the Gracenote API.  We use word shift graphs to characterize the meaning of chord categories as well as categories of lyrics.

We find that Major chords are associated with higher valence lyrics than Minor chords, consistent with the previous music perception studies that showed that Major chords evoke more positive emotional responses than Minor chords~\cite{crowder1984perception,kastner1990perception, crowder1991perception, hunter2010feelings}.  For an intuition regarding the magnitude of the difference, the mean valence of Minor chord lyrics is approximately $6.16$ (e.g. the valence of the word ``people'' in our sentiment dictionary), while the mean valence of Major chord lyrics is approximately $6.28$ (e.g. the valence of the word ``community'' in our sentiment dictionary).  Interestingly, we also uncover that three types of 7th chords --- Dominant 7ths, Major 7ths, and Minor 7ths --- have even higher valence than Major chords. This effect has not been deeply researched, except for one music perception study which reported that, in contrast to our findings, 7th chords evoke emotional responses intermediate in valence between Major and Minor chords~\cite{lahdelma2014single}.

Significant variation exists in the lyrics associated with different geographical regions, musical genres, and historical eras.  For example, musical genres demonstrate an intuitive ranking when ordered by mean valence, ranging from low-valence \emph{Punk} and \emph{Metal} genres to high-valence \emph{Religious} and \emph{60's Rock} genres. We also found that sentiment declined over time from the 1950's until the 2000's.  Both of these findings are consistent with the results of a previous study conducted using a different dataset and lexicon~\cite{dodds2010measuring}.  At the same time, we report a new finding that the trend in declining valence has reversed itself, with lyrics from the 2010's era having higher valence than those from the 2000's era.  Finally, we perform a analysis of the variation of valence among geographical regions. We find that  \emph{Asia} has the highest valence  while \emph{Scandinavia} has the lowest (likely due to prevalence of `dark' genres in that region).
 
We perform a novel data-driven analysis of the Major vs. Minor distinction by measuring the \emph{difference} between Major and Minor valence for different regions, genres, and eras. All examined genres except \emph{Contemporary R\&B/Soul}
exhibited higher Major chord valence than Minor chord.  Interestingly, the
largest differences of Major above Minor may indicate genres (\emph{Classic
R\&B/Soul}, \emph{Classic Country}, \emph{Religious}, \emph{60's Rock}) that
are more musically `traditional'.  In terms of historical periods, we find
that, unlike other eras, the 1980's era did not have a significant Major-Minor
valence difference.  This phenomenon calls for further investigation; one
possibility is that it may be related to an important period of musical
innovation in 1980's, which was recently reported in a data-driven study of musical
evolution~\cite{mauch2015evolution}.  Finally, analysis of geographic variation
indicates that songs from the \emph{Asian} region---unlike those from other
regions---do not show a significant difference in the valence of Major vs.
Minor chords. In fact, it is known in the musicological literature that the association of positive emotions with Major chords and negative emotions with Minor chords is culture-dependent, and that some Asian cultures do not display this
association~\cite{malm2000traditional}.  Our results may provide new supporting evidence of cultural variation in the emotional connotations of the Major/Minor distinction.

Finally, we evaluate how much of the variation in valence in our dataset is attributable to chord category, genre, era, and region (we call these four types of attributes `explanatory factors').  We find that genre is the most explanatory, followed by era, chord category, and region.  We use statistical model selection to evaluate whether certain explanatory factors `mediate' the influence of others (an example of mediation would be if variation in valence of different eras is actually due to variation in the prevalence of different genres during those eras).  We find that all four explanatory factors are important for explaining variation in valence; no explanatory factors totally mediate the effect of others.

Our approach has several limitations.  First, the accuracy of tablature chord annotations may be limited  because users are not generally professional musicians; for instance, more complex chords (e.g. \texttt{dim} or \texttt{11th}) may be mis-transcribed as simpler chords.  To deal with this, we analyze relatively basic chords---Major, Minor, and 7ths---and, when there are multiple versions of a song, use tabs with the highest user-assigned rating. Our manual inspection of a small sample of parsed tabs indicated acceptable quality, although a more systematic evaluation of the error rate can be performed using more extensive manual inspection of tabs by professional musicians. 

There are also significant biases in our dataset.  We only consider tablatures for songs  entered by users of \texttt{ultimate-guitar.com}, which is likely to be heavily biased towards North American and European users and songs. In addition, this dataset is restricted to songs  playable by guitar, which selects for guitar-oriented musical genres and may be biased toward emotional meanings tied to guitar-specific acoustic properties, such as the instrument's timbre. 
Furthermore, our dataset includes only English-language songs and is not likely to be a representative sample of popular music from non-English speaking regions. Thus, for example, the high valence of songs from \emph{Asia} should not be taken as conclusive evidence that Asian popular music is overall happier than popular music in English-speaking countries.  For this reason, the absence of a Major vs. Minor chord distinction in \emph{Asia} is speculative and requires significant further investigation.

Despite these limitations, we believe that our results reveal meaningful patterns of association between music and emotion at least in guitar-based English-language popular music, and show the potential of our novel data-driven methods. At the same time, applying these methods to other datasets --- in particular those
representative of other geographical regions, historical eras, instruments, and musical styles --- is of great interest for future work.  
Another promising direction for future work is to move the analysis of emotional content beyond single chords, since emotional meaning is likely to be more closely associated with melodies rather than individual chords.  For this reason, we hope to extend our methodology to study chord progressions.

\vskip1pc

\dataccess{The datasets generated during and/or analyzed during the current study are available in the \texttt{figshare} repository, \url{https://goo.gl/R9CqtH}.
Code for performing the analysis and generating plots in this manuscript is available at \url{https://github.com/artemyk/chordsentiment}.
}

\aucontribute{N.D. and Y.Y.A. conceived the idea of analyzing the association between lyric sentiment and chord categories, while A.K. contributed the idea of analyzing the variation of this association across genres, eras, and regions. N.D. and A.K. downloaded and processed the dataset. All authors analyzed the dataset and results. A.K. and N.D. prepared the initial draft of the manuscript. All authors reviewed and edited the manuscript.}

\competing{The author(s) declare no competing financial interests.}

\funding{Y.Y.A thanks Microsoft Research for MSR Faculty Fellowship.}

\ack{We would like to thank Rob Goldstone, Christopher Raphael, Peter Miksza, Daphne Tan, and Gretchen Horlacher for helpful discussions and comments.}

\pagebreak

\bibliographystyle{RS} 
\bibliography{main} 

\end{document}